# Bayesian Active Distance Metric Learning

**Liu Yang and Rong Jin**
Dept. of Computer Science and Engineering
Michigan State University
East Lansing, MI 48824

**Rahul Sukthankar**[*]
Robotics Institute
Carnegie Mellon University
Pittsburgh, PA 15213

## Abstract

Distance metric learning is an important component for many tasks, such as statistical classification and content-based image retrieval. Existing approaches for learning distance metrics from pairwise constraints typically suffer from two major problems. First, most algorithms only offer point estimation of the distance metric and can therefore be unreliable when the number of training examples is small. Second, since these algorithms generally select their training examples at random, they can be inefficient if labeling effort is limited. This paper presents a Bayesian framework for distance metric learning that estimates a posterior distribution for the distance metric from labeled pairwise constraints. We describe an efficient algorithm based on the variational method for the proposed Bayesian approach. Furthermore, we apply the proposed Bayesian framework to active distance metric learning by selecting those unlabeled example pairs with the greatest uncertainty in relative distance. Experiments in classification demonstrate that the proposed framework achieves higher classification accuracy and identifies more informative training examples than the non-Bayesian approach and state-of-the-art distance metric learning algorithms.

## 1 Introduction

Learning application-specific distance metrics from labeled data is critical for both statistical classification and information retrieval. The essential goal of distance metric learning is to identify an appropriate distance metric that brings "similar" objects close together while separating "dissimilar" objects. A number of algorithms (Friedman, 1994; Hastie & Tibshirani, 1996; Domeniconi & Gunopulos, 2002; Xing et al., 2003; Goldberger et al., 2005; Weinberger et al., 2006; Kwok & Tsang, 2003; Bar-Hillel et al., 2003; Hoi et al., 2006; Globerson & Roweis, 2006; Yang et al., 2006; Schultz & Joachims, 2004) have been developed for distance metric learning. However, most of them assume that the number of labeled examples is sufficiently large for learning a reliable distance metric. Given a limited number of labeled examples, most distance metric learning algorithms would suffer from the following two problems: (1) since most algorithms only provide point estimation of the distance metric that could be sensitive to the choice of training examples, they tend to be unreliable when the number of training examples is small. (2) since most algorithms only randomly select examples for manual labeling, they tend to be inefficient when the number of labeled examples is limited.

In order to address the above two problems, this paper presents a Bayesian framework for distance metric learning, termed **Bayesian Distance Metric Learning**, that targets tasks where the number of training examples is limited. Using the full Bayesian treatment, the proposed framework for distance metric learning is better suited to dealing with a small number of training examples than the non-Bayesian approach. Furthermore, the proposed framework estimates not only the most likely distance metric, but also the uncertainty (i.e., the posterior distribution) for the estimated distance metric, which is further used for **Active Distance Metric Learning**. The key idea behind active distance metric learning is to select those unlabeled example pairs with the largest uncertainty in their relative distance as candidates for manual labeling. Two efficient algorithms are presented in this study to facilitate the computation: a variational approach for Bayesian distance metric learning, and an active distance metric learning algorithm based on the Laplacian approximation. We demonstrate the efficacy of these two algorithms through the empirical studies with image classification and spoken letter recognition.

The remainder of the paper is organized as follows. Section 2 reviews the previous work on distance metric learn-ing and active learning. Section 3 outlines the Bayesian framework for distance metric learning. Section 4 describes an active learning algorithm using the Bayesian framework. Section 5 presents empirical results in image classification and spoke letter recognition. Section 6 concludes the paper.

---

[*]Intel Research Pittsburgh

## 2 Related Work

We first review previous work on distance metric learning, followed by an overview of active learning.

**Distance Metric Learning** Our work is closely related to the previous study on supervised distance metric learning. Most algorithms in this area learn a distance metric from side information that is typically presented in a set of pairwise constraints: equivalence constraints that include pairs of "similar" objects and inequivalence constraints that include pairs of "dissimilar" objects. The optimal distance metric is found by keeping objects in equivalence constraints close, and at the same time, objects in inequivalence constraints well separated. In the past, a number of algorithms have been developed for supervised distance metric learning. (Xing et al., 2003) formulates distance metric learning into a constrained convex programming problem by minimizing the distance between the data points in the same classes under the constraint that the data points from different classes are well separated. This algorithm is extended to the nonlinear case in (Kwok & Tsang, 2003) by the introduction of kernels. Local linear discriminative analysis (Hastie & Tibshirani, 1996) estimates a local distance metric using the local linear discriminant analysis. Relevant Components Analysis (RCA) (Bar-Hillel et al., 2003) learns a global linear transformation from the equivalence constraints. The learned linear transformation can be used directly to compute distance between any two examples. Discriminative Component Analysis (DCA) and Kernel DCA (Hoi et al., 2006) improve RCA by exploring negative constraints and aiming to capture nonlinear relationships using contextual information. (Schultz & Joachims, 2004) extends the support vector machine to distance metric learning by encoding the pairwise constraints into a set of linear inequalities. Neighborhood Component Analysis (NCA) (Goldberger et al., 2005) learns a distance metric by extending the nearest neighbor classifier. The maximum-margin nearest neighbor (LMNN) classifier (Weinberger et al., 2006) extends NCA through a maximum margin framework. (Globerson & Roweis, 2006) learns a Mahalanobis distance that tries to collapse examples in the same class to a single point, and in the meantime keep examples from different classes far away. (Yang et al., 2006) propose a Local Distance Metric (LDM) that addresses multimodal data distributions in distance metric learning by optimizing local compactness and local separability in a probabilistic framework. An comprehensive overview of distance metric learning can be found in (Yang & Jin, 2006).

Despite extensive development, most algorithms only provide point estimation for distance metric, which could be unreliable when the number of training examples is small. Furthermore, most previous work assumes randomly-selected training examples, which could be insufficient in identifying the optimal distance metric. The proposed framework aims to address these problems by a full Bayesian treatment and active distance metric learning.

**Active Learning** Previous work on active learning largely focuses on classification problems. The key idea behind most active learning algorithms is to select the examples that are most uncertain to classify. Therefore, a key aspect of active learning is to measure the classification uncertainty of unlabeled examples. In (Seung et al., 1992; Abe & Mamitsuka, 1998; Melville & Mooney, 2004), the authors propose to measure the classification uncertainty of an example by the disagreement in the class labels predicted by an ensemble of classification models. In (Tong & Koller, 2000b; Campbell et al., 2000; Roy & McCallum, 2001), the classification uncertainty of an example is measured by its distance to the decision boundary. (MacKay, 1992; Zhang & Oles, 2000) represent the uncertainty of a classification model by its Fisher information matrix, and measure the classification uncertainty of an unlabeled example by its projection onto the Fisher information matrix. In (Tong & Koller, 2000a; Zhang et al., 2003; Jin & Si, 2004; Freund et al., 1997), the Bayesian analysis is used for active learning that takes into account the model distribution. In addition, several other approaches (Hofmann & Buhmann, 1997; Cohn et al., 1995) are developed for active learning. The active learning work that is closely related to this study is (X. Zhu & Ghahramani, 2003) and (Sugato Basu & Mooney, 2004). Both studies aim to select the most informative example pairs. However, the goals of these two studies differ from that of this work: (X. Zhu & Ghahramani, 2003) examines the active learning problem for ordinal regression, and (Sugato Basu & Mooney, 2004) seeks the example pairs to effectively improve the accuracy of data clustering.

## 3 A Bayesian Framework for Distance Metric Learning

We first present the Bayesian framework for distance metric learning, followed by the efficient algorithm using the variational approximation.

### 3.1 The Bayesian Framework

To introduce the Bayesian framework for distance metric learning, we first define the probability for two data points $x_i$ and $x_j$ to form an equivalence or inequivalence con-straint under a given distance metric $\mathbf{A}$:

$$\Pr(y_{i,j}|\mathbf{x}_i, \mathbf{x}_j, \mathbf{A}, \mu) = \frac{1}{1 + \exp\left(y_{i,j}(\|\mathbf{x}_i - \mathbf{x}_j\|_\mathbf{A}^2 - \mu)\right)} \quad (1)$$

$$\text{where } y_{i,j} = \begin{cases} +1 & (\mathbf{x}_i, \mathbf{x}_j) \in \mathcal{S} \\ -1 & (\mathbf{x}_i, \mathbf{x}_j) \in \mathcal{D} \end{cases}$$

In the above, $\mathcal{S}$ and $\mathcal{D}$ denote the sets of equivalence and inequivalence constraints, respectively. Parameter $\mu$ in Equation (1) stands for the threshold. Two data points are more likely to be assigned to the same class only when their distance is less than the threshold $\mu$. Using the expression in Equation (1), the overall likelihood function for all the constraints in $\mathcal{S}$ and $\mathcal{D}$ is written as:

$$\begin{aligned}\Pr(\mathcal{S},\mathcal{D}|\mathbf{A},\mu) &= \prod_{(i,j)\in\mathcal{S}} \frac{1}{1 + \exp\left(\|\mathbf{x}_i - \mathbf{x}_j\|_\mathbf{A}^2 - \mu\right)} \quad (2)\\ &\times \prod_{(i,j)\in\mathcal{D}} \frac{1}{1 + \exp\left(-\|\mathbf{x}_i - \mathbf{x}_j\|_\mathbf{A}^2 + \mu\right)}\end{aligned}$$

Furthermore, we introduce a Wishart prior for the distance metric $\mathbf{A}$ and a Gamma prior for the threshold $\mu$, i.e.,

$$\Pr(\mathbf{A}) = \frac{|\mathbf{A}|^{(\nu-m-1)/2}}{Z_\nu(\mathbf{W})} \exp\left(-\frac{1}{2}\mathrm{tr}(\mathbf{W}^{-1}\mathbf{A})\right) \quad (3)$$

$$\Pr(\mu) = \frac{\mu^{\alpha-1}}{Z(\alpha)}\exp(-\beta\mu) \quad (4)$$

where $Z_\nu(\mathbf{W})$ and $Z(\alpha)$ are the normalization factors. By putting the priors and the likelihood function together, we can estimate the posterior distribution as follows:

$$\Pr(\mathbf{A},\mu|\mathcal{S},\mathcal{D}) = \frac{\Pr(\mathbf{A})\Pr(\mu)\Pr(\mathcal{S},\mathcal{D}|\mathbf{A})}{\int_{\mathbf{A}\in\mathbf{S}_+} d\mathbf{A}\,\Pr(\mathbf{A})\int_0^\infty d\mu\,\Pr(\mu)\Pr(\mathcal{S},\mathcal{D}|\mathbf{A},\mu)} \quad (5)$$

Estimating the posterior distribution $\Pr(\mathbf{A},\mu|\mathcal{S},\mathcal{D})$ using the above equation is computationally expensive because that it involves an integration over the space of positive semi-definitive matrices. In the next section, we introduce an efficient algorithm for computing $\Pr(\mathbf{A},\mu|\mathcal{S},\mathcal{D})$.

### 3.2 An Efficient Algorithm

The proposed efficient algorithm consists of two steps. First, we approximate the distance metric $\mathbf{A}$ as an linear combination of the top eigenvectors of the observed data. Second, we estimate the posterior distribution of the combination weights using a variational method.

#### 3.2.1 Eigen Approximation

To simplify the computation, we model the distance metric $\mathbf{A}$ as a simple parametric form by the top eigenvectors of observed data points. Specifically, let $\mathbf{X} = (\mathbf{x}_1, \mathbf{x}_2, \ldots, \mathbf{x}_n)$ denote all the examples, including both the labeled examples that are used by the constraints in $\mathcal{S}$ and $\mathcal{D}$, and the unlabeled examples [1]. Let $\mathbf{v}_i, i = 1, 2, \ldots, K$ be the top $K$ eigenvectors of $\mathbf{X}\mathbf{X}^\top$. We then assume $\mathbf{A} = \sum_{i=1}^K \gamma_i \mathbf{v}_i \mathbf{v}_i^\top$, where $\gamma_i \geq 0, i = 1, 2, \ldots, K$ are the combination weights. Using the above expression for $\mathbf{A}$, we can rewrite the likelihood $\Pr(y_{i,j}|\mathbf{x}_i, \mathbf{x}_j)$ in (1) as follows:

$$\begin{aligned}\Pr(y_{i,j}|\mathbf{x}_i, \mathbf{x}_j) &= \frac{1}{1 + \exp\left(y_{i,j}(\sum_{l=1}^K \gamma_l \omega_{i,j}^l - \mu)\right)}\\ &= \sigma(-y_{i,j}\gamma^\top \omega_{i,j}) \quad (6)\end{aligned}$$

where $\omega_{i,j}^l = [(\mathbf{x}_i - \mathbf{x}_j)^\top \mathbf{v}_l]^2$ and $\sigma(z) = 1/(1 + \exp(-z))$. Note that, in the above, to simplify our notation, we augment the vector $\gamma$ and $\omega$ as follows:

$$\begin{aligned}\gamma &= (\mu, \gamma_1, \gamma_2, \ldots, \gamma_K)\\ \omega_{i,j} &= (-1, \omega_{i,j}^1, \omega_{i,j}^2, \ldots, \omega_{i,j}^K).\end{aligned}$$

Using the above approximation, we reduce the Wishart prior in Equation (3) into the product of a number of Gamma distributions, i.e., $\Pr(\mathbf{A}) = \prod_{i=1}^K \mathcal{G}(\gamma_i; \alpha, \beta)$. For the sake of computational simplicity, we relax the above Gamma distributions to a set of Gaussian distributions, which leads to the following expression for the prior distribution:

$$\begin{aligned}\Pr(\mathbf{A})\Pr(\mu) &\approx \prod_{i=1}^{K+1} \mathcal{N}(\gamma_i; \gamma_0, \delta^{-1})\\ &= \mathcal{N}(\gamma; \gamma_0 \mathbf{1}_{K+1}, \delta^{-1} I_{K+1}) \quad (7)\end{aligned}$$

where $\mathcal{N}(\mathbf{x}; \mu, \Sigma)$ is the Gaussian distribution with mean as $\mu$ and covariance matrix as $\Sigma$. Evidently, one problem with the above relaxation is that combination weights $\gamma$ is no long guaranteed to be non-negative. This problem is solved empirically by enforcing the mean of the $\gamma$ to be non-negative. Finally, using the expressions in Equations (6) and (7), evidence $\Pr(\mathcal{S}, \mathcal{D})$ is computed as:

$$\begin{aligned}\Pr(\mathcal{S},\mathcal{D}) &= \int d\mathbf{A}\,\Pr(\mathbf{A})\int d\mu\,\Pr(\mu)\Pr(\mathcal{S},\mathcal{D}|\mathbf{A},\mu)\\ &\approx \int d\gamma\,\mathcal{N}(\gamma; \gamma_0 \mathbf{1}_{K+1}, \delta^{-1}I_{K+1})\\ &\quad \prod_{(\mathbf{x}_i,\mathbf{x}_j)\in\mathcal{S}} \sigma(-\gamma^\top \omega_{i,j}) \prod_{(\mathbf{x}_i,\mathbf{x}_j)\in\mathcal{D}} \sigma(\gamma^\top \omega_{i,j}) \quad (8)\end{aligned}$$

#### 3.2.2 Variational Approximation

As the second step of simplification, to estimate the posterior distribution for $\gamma$, we employ the variational method (Jordan et al., 1999). The main idea is to introduce variational distributions for $\gamma$s to construct the lower bound for the logarithm of the evidence, i.e., $\log \Pr(\mathcal{S}, \mathcal{D})$. By maximizing the variational distributions with respect to the lower bound, we obtain the approximate estimation for the posterior distribution of $\gamma$s. More specifically, given the variational distribution $\phi(\gamma)$, the logarithm of the evidence is lower bounded by the following expression:

$$\begin{aligned} & \log \Pr(\mathcal{S}, \mathcal{D}) \\ =\ & \log \int d\gamma \Pr(\gamma) \prod_{(i,j) \in \mathcal{S}} \Pr(+|\mathbf{x}_i, \mathbf{x}_j) \prod_{(i,j) \in \mathcal{D}} \Pr(-|\mathbf{x}_i, \mathbf{x}_j) \\ \geq\ & \langle \log \Pr(\gamma) \rangle + H(\phi(\gamma)) + \sum_{(i,j) \in \mathcal{S}} \langle \log \Pr(+|\mathbf{x}_i, \mathbf{x}_j) \rangle \\ & + \sum_{(i,j) \in \mathcal{D}} \langle \log \Pr(-|\mathbf{x}_i, \mathbf{x}_j) \rangle. \end{aligned}$$

where $\langle \cdot \rangle = \langle \cdot \rangle_{\phi_\gamma}$. Using the inequality (Jaakkola & Jordan, 2000)

$$\sigma(z) \geq \sigma(\xi) \exp\left( \frac{z-\xi}{2} - \lambda(\xi)(z^2 - \xi^2) \right),$$

where $\lambda(\xi) = \tanh(\xi/2)/(4\xi)$, we can lower bound $\langle \log \Pr(y|\mathbf{x}_i, \mathbf{x}_j) \rangle$ by the following expression:

$$\begin{aligned} \langle \log \Pr(y|\mathbf{x}_i, \mathbf{x}_j) \rangle \geq\ & \log \sigma(\xi_{i,j}) + \frac{-y \langle \gamma \rangle^\top \omega_{i,j} - \xi_{i,j}}{2} \\ & - \lambda(\xi_{i,j}) \left( \mathrm{tr}(\omega_{i,j} \omega_{i,j}^\top \langle \gamma \gamma^\top \rangle) - \xi_{i,j}^2 \right). \end{aligned}$$

---

[1] Note the unlabeled samples used to compute eigen approximation are not necessarily testing samples.

Now using the above expression for the lower bound, we obtain a new expression to bound the evidence function: $\log \Pr(\mathcal{S}, \mathcal{D})$, i.e.,

$$\begin{aligned} \log \Pr(\mathcal{S}, \mathcal{D}) \geq\ & \langle \log \Pr(\gamma) \rangle + H(\phi(\gamma)) \\ & + \sum_{(i,j) \in \mathcal{S}} \left( \log(\sigma(\xi_{i,j}^s)) - \frac{\langle \gamma \rangle^\top \omega_{i,j}^s + \xi_{i,j}^s}{2} \right) \\ & + \sum_{(i,j) \in \mathcal{D}} \left( \log(\sigma(\xi_{i,j}^d)) + \frac{\langle \gamma \rangle^\top \omega_{i,j}^d - \xi_{i,j}^d}{2} \right) \\ & - \sum_{(i,j) \in \mathcal{S}} \lambda(\xi_{i,j}^s) \left( \mathrm{tr}(\omega_{i,j}^s [\omega_{i,j}^s]^\top \langle \gamma \gamma^\top \rangle) - [\xi_{i,j}^s]^2 \right) \\ & - \sum_{(i,j) \in \mathcal{D}} \lambda(\xi_{i,j}^d) \left( \mathrm{tr}(\omega_{i,j}^d [\omega_{i,j}^d]^\top \langle \gamma \gamma^\top \rangle) - [\xi_{i,j}^d]^2 \right). \end{aligned}$$

In above, we introduce variational parameters $\xi_{i,j}^s$ and $\xi_{i,j}^d$ for every pairwise constraint in $\mathcal{S}$ and $\mathcal{D}$, respectively. By maximizing the posterior distribution $\phi(\gamma)$ with respect to the lower bound of the evidence function, we have $\phi(\gamma) \sim \mathcal{N}(\gamma; \mu_\gamma, \Sigma_\gamma)$ where the mean $\mu_\gamma$ and the covariance matrix $\Sigma_\gamma$ are computed by the following update equations:

$$\Sigma_\gamma = (\delta I_K + 2\Sigma_\mathcal{S} + 2\Sigma_\mathcal{D})^{-1} \quad (9)$$

$$\mu_\gamma = \Sigma_\gamma \left( \delta \gamma_0 - \sum_{(i,j) \in \mathcal{S}} \frac{\omega_{i,j}^s}{2} + \sum_{(i,j) \in \mathcal{D}} \frac{\omega_{i,j}^d}{2} \right) \quad (10)$$

where $\Sigma_\mathcal{S}$ and $\Sigma_\mathcal{D}$ are defined as follows:

$$\begin{aligned} \Sigma_\mathcal{S} &= \sum_{(\mathbf{x}_i, \mathbf{x}_j) \in \mathcal{S}} \lambda(\xi_{i,j}^s) \omega_{i,j}^s [\omega_{i,j}^s]^\top, \\ \Sigma_\mathcal{D} &= \sum_{(\mathbf{x}_i, \mathbf{x}_j) \in \mathcal{D}} \lambda(\xi_{i,j}^d) \omega_{i,j}^d [\omega_{i,j}^d]^\top. \end{aligned}$$

Finally, according to (Jaakkola & Jordan, 2000), the variational parameters $\xi_{i,j}^s$ and $\xi_{i,j}^d$ are estimated as follows:

$$\begin{aligned} \xi_{i,j}^s &= \sqrt{[\mu_\gamma^\top \omega_{i,j}^s]^2 + [\omega_{i,j}^s]^\top \Sigma_\gamma \omega_{i,j}^s}, \\ \xi_{i,j}^d &= \sqrt{[\mu_\gamma^\top \omega_{i,j}^d]^2 + [\omega_{i,j}^d]^\top \Sigma_\gamma \omega_{i,j}^d}. \quad (11) \end{aligned}$$

Based on the above derivation, the combination weights $\gamma$s are updated through EM-like iterations. In the E-step, given the values for the variational parameters $\xi_{i,j}^d$ and $\xi_{i,j}^s$, we compute the mean $\mu_\gamma$ and the covariance matrix $\Sigma_\gamma$ using Equations (10) and (9). In the M-step, we recompute the optimal value for $\xi_{i,j}^d$ and $\xi_{i,j}^s$ using Equation (11) based on the estimated mean $\mu_\gamma$ and covariance matrix $\Sigma_\gamma$.

## 4 Bayesian Active Distance Metric Learning

To select data pairs that are informative for the target distance metric, we follow the uncertainty principle for active learning. In particular, we will select the pair of data points with the largest uncertainty in deciding whether or not the two data points are close to each other. For a given data pair $(\mathbf{x}_i, \mathbf{x}_j)$, this uncertainty is measured by the following entropy function $H_{i,j}$

$$\begin{aligned} H_{i,j} =\ & - \Pr(-|\mathbf{x}_i, \mathbf{x}_j) \log \Pr(-|\mathbf{x}_i, \mathbf{x}_j) \\ & - \Pr(+|\mathbf{x}_i, \mathbf{x}_j) \log \Pr(+|\mathbf{x}_i, \mathbf{x}_j). \quad (12) \end{aligned}$$

Thus, the key question is how to efficiently compute $\Pr(\pm|\mathbf{x}_i, \mathbf{x}_j)$. In the simplest form, $\Pr(\pm|\mathbf{x}_i, \mathbf{x}_j)$ can be directly computed using $\gamma$, i.e.,

$$\Pr(\pm|\mathbf{x}_i, \mathbf{x}_j) = \frac{1}{1 + \exp(\pm \gamma^\top \omega_{i,j})} \quad (13)$$

Since the entropy function $H_{i,j}$ in (12) is a monotonically decreasing function in $|\Pr(+|\mathbf{x}_i, \mathbf{x}_j) - \frac{1}{2}|$, it is therefore monotonically decreasing in $|\gamma^\top \omega_{i,j}|$. We thus can simply compute the quality $|\gamma^\top \omega_{i,j}|$ to indicate the uncertainty in labeling the example pair $(\mathbf{x}_i, \mathbf{x}_j)$.

However, the above computation does not take into account the distribution of $\gamma$. To incorporate the full distribution of $\gamma$, we can compute the conditional probability $\Pr(\pm|\mathbf{x}_i, \mathbf{x}_j)$ computed as follows:

$$\begin{aligned} \Pr(\pm|\mathbf{x}_i, \mathbf{x}_j) &= \int \frac{\mathcal{N}(\gamma; \Sigma_\gamma, \mu_\gamma)}{1 + \exp(\pm \gamma^\top \omega_{i,j})} d\gamma \\ &\propto \int \exp\left( -l_{i,j}^\pm(\gamma) \right) d\gamma \end{aligned}$$

where $l_{i,j}^{\pm}(\gamma)$ is defined as

$$\begin{aligned} l_{i,j}^{\pm}(\gamma) &= \log\left(1 + \exp(\pm\gamma^\top \omega_{i,j})\right) \\ &+ \frac{1}{2}(\gamma - \mu_\gamma)^\top \Sigma_\gamma^{-1}(\gamma - \mu_\gamma). \end{aligned}$$

To effectively evaluate the probability, we employ the Laplacian approximation. In particular, we first approximate $l_{i,j}^{\pm}(\gamma)$ by its Taylor expansion around the optimal point $\gamma_{i,j}^{\pm}$. We then compute the integral using the approximated $l_{i,j}^{\pm}(\gamma)$. However, this involves solving the optimization $\gamma_{i,j}^{\pm} = \arg\min_{\gamma \geq 0} l_{i,j}^{\pm}(\gamma)$ for each data pair, which is computationally expensive when the number of candidate data pairs is large. To further simplify the computation, we approximate the optimal solution $\gamma_{i,j}^{\pm}$ by expanding $l_{i,j}^{\pm}(\gamma)$ in the neighborhood of $\mu_\gamma$ as follows:

$$\begin{aligned} l_{i,j}^{\pm}(\gamma) &\approx \log(1 + \exp(\pm\mu_\gamma^\top \omega_{i,j})) \pm p_{i,j}^{\pm}(\gamma - \mu_\gamma)^\top \omega_{i,j} \\ &+ \frac{1}{2}(\gamma - \mu_\gamma)^\top \left(\Sigma_\gamma^{-1} + q_{i,j}\omega_{i,j}\omega_{i,j}^\top\right)(\gamma - \mu_\gamma) \\ &\approx \log(1 + \exp(\pm\mu_\gamma^\top \omega_{i,j})) + p_{i,j}^{\pm}(\gamma - \mu_\gamma)^\top \omega_{i,j} \\ &+ \frac{1}{2}(\gamma - \mu_\gamma)^\top \Sigma_\gamma^{-1}(\gamma - \mu_\gamma) \end{aligned}$$

where

$$p_{i,j}^{\pm} = \frac{\exp(\pm\mu_\gamma^\top \omega_{i,j})}{1 + \exp(\pm\mu_\gamma^\top \omega_{i,j})}, \quad q_{i,j} = p_{i,j}(1 - p_{i,j}) \quad (14)$$

In the above, we approximate $(\Sigma_\gamma^{-1} + q_{i,j}\omega_{i,j}\omega_{i,j}^\top)$ as $\Sigma_\gamma^{-1}$. This is because according to (9), $\Sigma_\gamma^{-1}$ is a summation across all the labeled example pairs, and therefore is significantly more important than the single term $q_{i,j}\omega_{i,j}\omega_{i,j}^\top$. Then, the approximate solutions for $\gamma_{i,j}^{\pm}$ and $l_{i,j}^{\pm}(\gamma)$ are

$$\gamma_{i,j}^{\pm} \approx \max(\mu_\gamma \mp p_{i,j}^{\pm}\Sigma_\gamma\omega_{i,j}, \mathbf{0}) \quad (15)$$

$$l_{i,j}^{\pm}(\gamma) \approx l_{i,j}^{\pm}(\gamma_{i,j}^{\pm}) + \frac{(\gamma - \gamma_{i,j}^{\pm})^\top \Sigma_\gamma^{-1}(\gamma - \gamma_{i,j}^{\pm})}{2} \quad (16)$$

The max operator in the above refers to element wise maximization. As indicated in (15), the optimal $\gamma^{\pm}$ is determined by both the mean $\mu_\gamma$ and the covariance matrix $\Sigma_\gamma$. Furthermore, the posterior $\Pr(\pm|\mathbf{x}_i, \mathbf{x}_j)$ is computed as:

$$\begin{aligned} \Pr(\pm|\mathbf{x}_i, \mathbf{x}_j) &\propto \exp\left(-l_{i,j}^{\pm}(\gamma_{i,j}^{\pm})\right) \\ &= \frac{1}{1 + \exp(\pm\omega_{i,j}^\top \gamma_{i,j}^{\pm})} \exp\left(-\frac{[p_{i,j}^{\pm}]^2 \omega_{i,j}\Sigma_\gamma\omega_{i,j}}{2}\right) \end{aligned}$$

Note that the estimation of probability $\Pr(\pm|\mathbf{x}_i, \mathbf{x}_j)$ in the above expression takes into account both the mean and the variance of the distribution of $\gamma$. Finally, $\Pr(\pm|\mathbf{x}_i, \mathbf{x}_j)$ are normalized to ensure $\Pr(+|\mathbf{x}_i, \mathbf{x}_j) + \Pr(-|\mathbf{x}_i, \mathbf{x}_j) = 1$.

## 5 Evaluation

In this section, we evaluate the proposed Bayesian framework for distance metric learning in the context of data classification. In particular, we will address the following two questions in this empirical study:

- *Is the proposed Bayesian framework effective for distance metric learning given a small number of labeled pairs?*
- *Is the proposed Bayesian framework effective for active distance metric learning?*

### 5.1 Experiment Methodology

Two datasets are used as our testbed:

- *COREL image database ("corel")*: We choose five categories from the COREL image database, and randomly select 100 images for each category, which amounts to a total of 500 images. Each image is represented by 36 different visual features from three different categories, i.e., color, edge and texture.
- *Spoken Letter Recognition ("isolet")*: This dataset comes from the standard UCI machine learning repository (Newman et al., 1998). It consists of 7,797 examples that belong to 26 classes. We select 10 classes, and 100 samples for each class to create a dataset with 1000 examples. Each example is originally represented by 617 features. We employ PCA to reduce the total number of dimensionality to 200.

The quality of the learned distance metric is evaluated by the Nearest Neighbor (1NN) classifier. More specifically, the learned distance metric is used to measure distance between the training examples and the test examples. For each test example, the class of the closest training example is predicted as the class label for the test example. Each experiment is repeated ten times, and both the mean and the standard deviation of the classification accuracy are reported in this study.

### 5.2 Distance Metric Learning with a Small Number of Training Examples

Our first experiment is to demonstrate that the proposed Bayesian framework is effective for distance metric learning when the number of training examples is small. As a direct comparison, we compare the Bayesian framework to a maximum likelihood based approach for distance metric learning. Specifically, this approach finds the optimal distance metric by maximizing the following likelihood function $\mathcal{L}_g(\gamma)$:

$$\begin{aligned} \mathcal{L}_g(\gamma) &= -\sum_{(i,j)\in\mathcal{S}} \log(1 + \exp(y_{i,j}\gamma^\top \omega_{i,j})) \\ &- \sum_{(i,j)\in\mathcal{D}} \log(1 + \exp(-y_{i,j}\gamma^\top \omega_{i,j})). \end{aligned}$$

We refer to this approach as **MLE**, and the Bayesian framework as **BAYES**. In addition, we compare the Bayesian approach to two state-of-the-art distance metric learning algorithms, i.e., Neighborhood Component Analysis (**NCA**) (Goldberger et al., 2005), and Maximum Margin Nearest Neighbor Classifier (**LMMN**) (Weinberger et al., 2006). The Euclidean distance, termed as **EUCLID**, is used as the reference point in this study.

Table 1: Classification accuracy of 1NN on the corel testbed using different distance metric learning algorithms.

| Training Size | **EUCLID** | **MLE** | **NCA** | **LMNN** | **BAYES** |
|---|---|---|---|---|---|
| 10 | $0.449 \pm 0.028$ | $0.439 \pm 0.026$ | $0.484 \pm 0.027$ | $0.487 \pm 0.026$ | $0.492 \pm 0.024$ |
| 20 | $0.547 \pm 0.019$ | $0.602 \pm 0.009$ | $0.603 \pm 0.021$ | $0.581 \pm 0.024$ | $0.610 \pm 0.020$ |
| 30 | $0.572 \pm 0.011$ | $0.596 \pm 0.012$ | $0.624 \pm 0.015$ | $0.619 \pm 0.023$ | $0.646 \pm 0.015$ |

Table 2: Classification accuracy of 1NN on the isolet testbed using different distance metric learning algorithms.

| Training Size | **EUCLID** | **MLE** | **NCA** | **LMNN** | **BAYES** |
|---|---|---|---|---|---|
| 10 | $0.608 \pm 0.034$ | $0.659 \pm 0.023$ | $0.663 \pm 0.020$ | $0.674 \pm 0.023$ | $0.681 \pm 0.018$ |
| 20 | $0.648 \pm 0.018$ | $0.695 \pm 0.012$ | $0.717 \pm 0.017$ | $0.728 \pm 0.021$ | $0.731 \pm 0.013$ |
| 30 | $0.704 \pm 0.019$ | $0.733 \pm 0.013$ | $0.743 \pm 0.013$ | $0.761 \pm 0.015$ | $0.770 \pm 0.020$ |

Some of the above algorithms for distance metric learning require labeled examples, not labeled example pairs. Therefore, for both datasets, we randomly select 10, 20, and 30 examples (not example pairs) for learning distance metric. To ensure the diversity of the training data, the same number of training examples is selected for each class. 100 examples are randomly selected for testing and the remaining examples are used for training 1NN.

Table 1 and 2 summarize the classification accuracies of 1NN for the corel and the isolet testbed using the distance metrics that are learned by different algorithms, respectively. First, we observe that all the distance metric learning algorithms are able to outperform the Euclidean distance metric significantly for both datasets except for the case when using the MLE approach with 10 training examples for the corel dataset. Second, compared to the MLE approach, we observe that the Bayesian approach is significantly more effective in improving the classification accuracy of 1NN classifier for all the cases. A t-test shows our performance gain is statistically significant at a significance level of $0.05$. It is interesting to notice that when the number of training examples is 10, the MLE approach performs slightly worse than the Euclidean approach for the corel testbed. This result clearly indicates the importance of employing full Bayesian treatment for distance metric learning when the number of training examples is small. Third, we find that the Bayesian approach perform slightly better than the two state-of-the-art distance metric learning approaches for both datasets with different number of training examples. Based on the above observation, we conclude that the proposed Bayesian approach is effective for distance metric learning, particularly when the number of training examples is small.

### 5.3 Active Distance Metric Learning

In this experiment, we evaluate the effectiveness of the proposed method for active distance metric learning. Our framework of Bayesian active distance metric learning, termed as **BAYES+VAR**, takes into account the distribution of $\gamma$ (i.e., both the mean and the covariance matrix) when evaluating the posterior probability $\Pr(\pm|\mathbf{x}_i, \mathbf{x}_j)$ for active pair selection. Three baseline approaches for active disance metric learning are used in this study:

- **BAYES+ACT**. This approach first applies the Bayesian distance metric learning algorithm to find the appropriate distance metric for the given training pairs. It then evaluates the posterior probability $\Pr(\pm|\mathbf{x}_i, \mathbf{x}_j)$ by directly using the mean $\mu_\gamma$. The example pairs with the largest entropy will be selected for manually labeling.
- **MLE+ACT**. This approach relies on the MLE approach presented in previous subsection to learn a distance metric from the training example pairs. Similar to the BAYES+ACT approach, it evaluates the posterior probability $\Pr(\pm|\mathbf{x}_i, \mathbf{x}_j)$ with the learned distance metric, and finds the most informative example pairs based on their entropy.
- **MLE+RAND**. This approach is similar to the MLE+ACT approach except that the example pairs are randomly selected for manually labeling.

Similar to the setup of the previous experiment, for both testbeds, 100 examples are randomly selected for testing, and the rest examples serve as the training data for the 1NN classifier. In addition, for both datasets, we randomly select 50 examples, with equal number of examples from each class, to form the pool of example pairs for training distance metrics. This results in a pool of $1,225$ training pairs. First, at the beginning of active distance metric learning, we randomly select 10 example pairs from the pool of training pairs to train a distance metric. Then, at each iteration of active learning, additional 20 example pairs are selected from the pool to train a new distance metric. The classification accuracy of 1NN is used to evaluate the quality of distance metric. Finally, we also run the experiments with 20 and 30 training pairs that are selected at the beginning of active learning.

Figure 1 and 2 show the classification accuracy of 1NN for the corel and the isolet testbed across different iterations of active distance metric learning. First, compared to the MLE+RAND algorithm, we observe that its active learning version, i.e., MLE+ACT, is significantly more effective in improving the classification accuracy of 1NN across all the iterations. This indicates that active learning approach is more effective in identifying appropriate distance metric than randomly selecting example pairs. Second, for the first few iterations, we observe a significant gap in classification accuracy of 1NN between the MLE approaches and the Bayesian approaches for distance metric learning. The performance gap starts to become smaller when more and more example pairs are added. This again indicates that the advantage of the Bayesian approach for distance metric learning arises when the number of training examples is small. Finally, compared to the BAYES+ACT approach that only utilizes the mean of $\gamma$ for evaluating the posterior probability $\Pr(\pm|\mathbf{x}_i, \mathbf{x}_j)$, the BAYES+VAR approach appears to be more effective in improving the classification accuracy of 1NN. This is demonstrated more clearly in the corel testbed where BAYES+VAR outperforms BAYES+ACT significantly across all the iterations with different number of training examples. The results for isolet testbed also shows us a clear advantage of BAYES+VAR compared to BAYES+ACT, when the number of initial training example pairs is 10 and 20. When the number of initial training example pairs is increased to 30, we start to observe a flat performance across all the iterations for both BAYES+VAR and BAYES+ACT. We will further investigate this issue in our future study.

## 6   Conclusion

This paper introduces a Bayesian framework for distance metric learning from labeled examples that computes the posterior distribution of a distance metric. In addition to the general framework, we also presented an efficient algorithm that exploits the variational method. Furthermore, we extend the Bayesian framework to active distance metric learning by selecting the example pairs with the largest uncertainty in their relative distance. An algorithm based on the Laplacian approximation was proposed to efficiently evaluate the posterior probability that decides if two examples should belong to the same class. Empirical evaluations on classification demonstrate that our proposed Bayesian approach clearly outperforms existing state-of-the-art methods. Our study also shows that the active distance metric learning based on the Bayesian framework is more effective than active learning using either random selection or maximum likelihood estimation.

## 7   Acknowledgments

This work is supported by the NSF grant IIS-0643494 and the NIH grant 1R01GM079688-01.

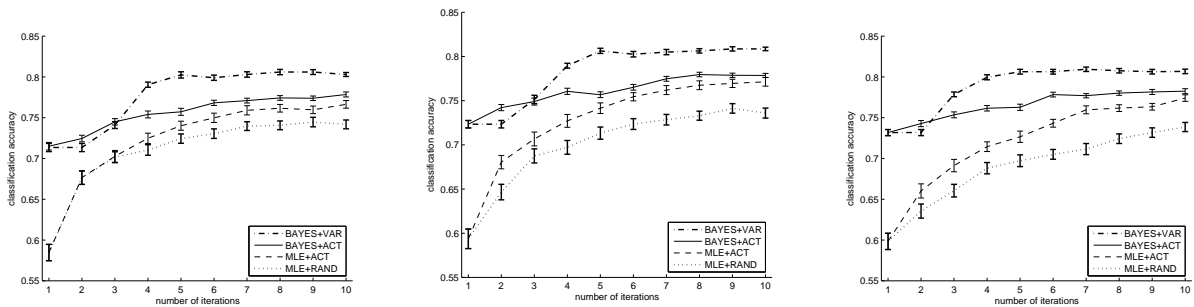

(a) #Initial training example pairs = 10    (b) #Initial training example pairs = 20    (c) #Initial training example pairs = 30

Figure 1: Classification accuracy of 1NN on the corel dataset for evaluating active distance metric learning.

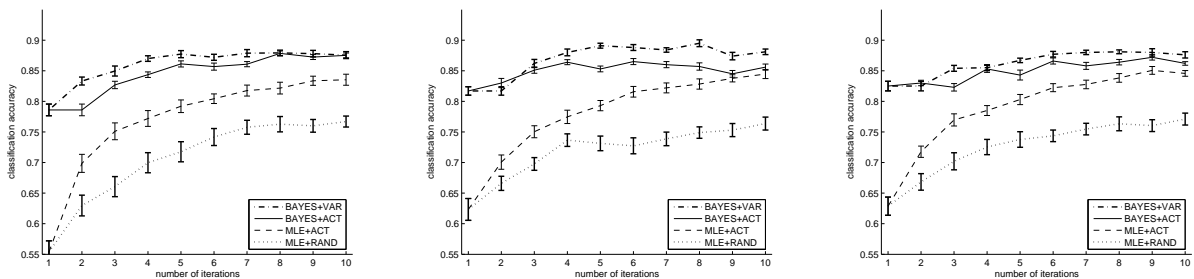

(a) #Initial training example pairs = 10    (b) #Initial training example pairs = 20    (c) #Initial training example pairs = 30

Figure 2: Classification accuracy of 1NN on the isolet dataset for evaluating active distance metric learning.